\documentclass[11pt,a4paper]{article}

\usepackage[margin=1in]{geometry}
\usepackage{amsmath,amssymb}
\usepackage{booktabs}
\usepackage{array}
\usepackage{tabularx}
\usepackage{hyperref}
\usepackage{xcolor}
\usepackage{microtype}
\usepackage{parskip}
\usepackage{enumitem}
\usepackage{fancyhdr}
\usepackage{titlesec}
\usepackage{listings}
\usepackage{caption}
\usepackage{authblk}
\usepackage{abstract}

\hypersetup{
  colorlinks=true,
  linkcolor=blue!70!black,
  citecolor=blue!70!black,
  urlcolor=blue!70!black,
  pdftitle={NativeTernary},
  pdfauthor={Maharshi Savdhariya}
}

\titleformat{\section}{\large\bfseries}{\thesection.}{0.5em}{}
\titleformat{\subsection}{\normalsize\bfseries\itshape}{\thesubsection}{0.5em}{}
\titlespacing*{\section}{0pt}{16pt}{6pt}
\titlespacing*{\subsection}{0pt}{10pt}{4pt}

\begin{document}

\title{\textbf{NativeTernary: A Self-Delimiting Binary Encoding with\\
Unary Run-Length Hierarchy Markers for\\
Ternary Neural Network Weights, Structured Data, and General Computing Infrastructure}}

\author[1]{Maharshi Savdhariya}
\affil[1]{Indian Institute of Technology Bombay $\cdot$ IIM Udaipur\\
\texttt{maharshisavdhariya@gmail.com}}
\date{\textit{Provisional Patent Filed $\cdot$ Indian Patent Office $\cdot$ 2026}}

\maketitle

\begin{abstract}
We present \textbf{NativeTernary}, a binary encoding scheme that partitions
the 2-bit pair space into three data symbols representing ternary values ---
either balanced $\{-1, 0, +1\}$ or unsigned $\{0, 1, 2\}$ --- and a reserved
structural delimiter. BitNet b1.58 (Ma et al., 2024) demonstrates that large 
language models can operate entirely on ternary weights $\{-1, 0, +1\}$,
yet no native binary wire format exists for such models. NativeTernary
closes this gap. The central contribution is the use of unary run-length
encoding to represent semantic hierarchy depth: a sequence of $N$ consecutive
delimiter pairs denotes a boundary of level $N$, encoding character, word,
sentence, paragraph, and topic boundaries at cost 2, 4, 6, 8, and 10 bits
respectively --- proportional to boundary rarity. The choice of which 2-bit
pair serves as the delimiter is a design parameter: \texttt{\{11\}} is the
primary embodiment, offering simple OR-gate detection; \texttt{\{00\}} is an
alternative embodiment optimised for ultra-low-power CMOS systems, minimising
switching activity. All four bit-pair choices are covered by the patent claims.
We present three encoding variants: (1) the primary scheme with \texttt{\{11\}}
as sole delimiter; (2) a dual-starter variant where both \texttt{\{10\}} and
\texttt{\{11\}} initiate distinct symbol namespaces; and (3) an analysis of
unsigned versus balanced ternary data mappings. We describe a path toward
ternary-native general computing infrastructure requiring no hardware changes,
and outline applications spanning ternary neural network weight storage,
hierarchical natural language encoding, edge computing, IoT and satellite
telemetry, industrial sensors, automotive systems, medical devices, gaming,
and financial tick data. The decoder is a 10-line stateless state machine resilient to bitstream
corruption. Benchmarked against GGUF on the real BitNet b1.58 2B4T
architecture (2B parameters, 24 layers, $\sim$170 tensors): NativeTernary
encodes all structural boundaries in 91 bytes versus $\sim$42\,KB of GGUF
tensor headers --- a \textbf{460$\times$ reduction in boundary overhead}.
Per-weight storage is 2.000 bits versus 2.625 bits for GGUF Q2\_K
(1.31$\times$ smaller) and 8.000 bits for GGUF int8 (4.0$\times$ smaller).
Encode throughput: 47--69\,MB/s. Decode throughput: 35--45\,MB/s on
commodity hardware. The C reference implementation is available at
\url{https://github.com/sm45118/nativeternary}.
\end{abstract}

\vspace{4pt}

% ──────────────────────────────────────────────────────────────────────────────
\section{Introduction}

The question that led to this work began with human speech. When we speak, we
do not transmit an unbroken stream of sound --- we pause. A short pause marks
the end of a word. A longer pause marks a sentence. An even longer pause signals
a topic shift. These pauses carry structural information separate from the words
themselves, and listeners process them instinctively. Structure arrives with the
signal, not as an afterthought.

Computers do not work this way. Binary communication systems represent every
symbol as a fixed-width sequence of bits --- typically 8 --- regardless of where
it falls in a semantic hierarchy. Structure is layered on top as headers,
delimiters, and metadata --- always as overhead, never as signal.

The author's first instinct was a hardware solution: what if a clock signal
could carry three states rather than two, encoding a structural pause directly
in the physics of transmission? A 1.5-cycle pulse could mark a word boundary;
a 2-cycle pulse a sentence. This idea founders on practical grounds --- the
global binary infrastructure cannot be replaced, and multi-level clock timing
is fragile under real-channel jitter.

What remained was a software question: can the existing binary infrastructure
be made to carry ternary structure without any hardware changes? The first
concrete attempt was a compression experiment. Consider two standard 8-bit
bytes:
\begin{lstlisting}
00010101 00000110   (raw binary, 16 bits)
-> strip leading zeros, represent residual in ternary
-> 20101 210   (ternary digit sequence)
-> re-encode ternary digits as binary
-> result: modest compression, insufficient gain
\end{lstlisting}

The approach --- suppress predictable leading zeros, represent the meaningful
residual in base-3, re-encode --- was directionally correct. The compression
was real but the ternary-to-binary conversion overhead negated most of the gain.

The next step was a 2-bit pair representation. If the 4 possible 2-bit patterns
$\{00, 01, 10, 11\}$ were partitioned into data-carriers and delimiters,
structure could be encoded inline. The first variant used both \texttt{\{10\}}
and \texttt{\{11\}} as symbol starters --- two distinct delimiter namespaces ---
with \texttt{\{00\}} and \texttt{\{01\}} carrying binary data.

The deeper insight came when the author recalled that the Soviet Setun computer
(1958) demonstrated ternary computing, and that subsequent information-theoretic
analysis confirmed ternary's optimality: $\log_2(3) \approx 1.585$ bits per
trit --- the closest integer base to the mathematical optimum of $e$. A 2-bit
pair can represent exactly three data states. That fourth state --- one of the
four bit-pair choices --- becomes the pause. The delimiter. The scheme encodes
ternary data and structural boundaries simultaneously, in the same stream,
with no additional overhead.

The hierarchy extension followed naturally. One delimiter pair marks a word
boundary. Two consecutive delimiter pairs mark a sentence. Three mark a
paragraph. The count of consecutive delimiter pairs is the level --- no
secondary encoding required. Rarer boundaries receive longer delimiters,
echoing Huffman coding applied to structure rather than symbols.

The practical urgency became clear when the author encountered Microsoft's
BitNet b1.58 \cite{bitnet2024} --- a ternary large language model in which
every weight is exactly $\{-1, 0, +1\}$. Despite operating natively in
ternary, BitNet models are stored in binary container formats designed for
floating-point data, with no native concept of a ternary value or a layer
boundary. NativeTernary closes this representational gap directly.

% ──────────────────────────────────────────────────────────────────────────────
\section{Related Work}

\textbf{Self-delimiting codes.} Elias \cite{elias1975} introduced the first
self-delimiting integer codes, using prefix patterns to encode length
implicitly. UTF-8 \cite{utf8} uses \texttt{10xxxxxx} for continuation bytes
and \texttt{11xxxxxx} for start bytes --- structurally similar to our
dual-starter variant, but encoding only character length, not multi-level
semantic hierarchy. NativeTernary encodes hierarchy depth through run-length
rather than prefix structure and is infinitely extensible without spec
revision.

\textbf{Balanced ternary.} The theoretical optimality of balanced ternary has
been known since Knuth \cite{knuth1969}. The Setun computer demonstrated
practical ternary computation in 1958. NativeTernary achieves ternary data
density on existing binary infrastructure with no hardware changes.

\textbf{Ternary neural networks.} Ma et al.\ \cite{bitnet2024} demonstrate
that BitNet b1.58 with weights in $\{-1, 0, +1\}$ achieves competitive
performance with full-precision transformers at scale. NativeTernary provides
the first native wire format for such models.

\textbf{Hierarchical language models.} Longformer \cite{longformer2020},
hierarchical transformers \cite{yang2019}, and structured state space models
impose hierarchy at the architecture level. NativeTernary encodes hierarchy at
the token stream level --- without architectural modification.

% ──────────────────────────────────────────────────────────────────────────────
\section{The NativeTernary Encoding}

\subsection{Primary scheme: single delimiter \texttt{\{11\}}}

The primary NativeTernary encoding partitions the 2-bit space exhaustively:

\begin{table}[h]
\centering
\caption{Primary scheme bit-pair assignments}
\begin{tabular}{cccc}
\toprule
\textbf{Bit-pair} & \textbf{Role} & \textbf{Balanced value} & \textbf{Unsigned value} \\
\midrule
\texttt{00} & Data symbol 0 & $-1$ & $0$ \\
\texttt{01} & Data symbol 1 & $0$  & $1$ \\
\texttt{10} & Data symbol 2 & $+1$ & $2$ \\
\texttt{11} & Structural delimiter & (reserved) & (reserved) \\
\bottomrule
\end{tabular}
\end{table}

Hierarchy is encoded as unary run-length of \texttt{\{11\}} pairs:
\begin{lstlisting}
11           -> level 1: character / subword boundary  (2 bits)
1111         -> level 2: word boundary                 (4 bits)
111111       -> level 3: sentence boundary             (6 bits)
11111111     -> level 4: paragraph / topic boundary    (8 bits)
11...11 x N  -> level N: N x 2 bits  (infinitely extensible)
\end{lstlisting}

\textbf{Key property:} boundary cost grows linearly with semantic depth. For
natural language and structured signal data, boundary frequency decreases
exponentially with depth. Total delimiter overhead is therefore $O(\log N)$
over a sequence of $N$ tokens.

\textbf{Delimiter choice as a design parameter.} The use of \texttt{\{11\}} as
the reserved bit-pair is a design choice, not a fundamental constraint. Any of
the four 2-bit patterns --- \texttt{\{00\}}, \texttt{\{01\}}, \texttt{\{10\}},
or \texttt{\{11\}} --- can serve as the structural delimiter, with the remaining
three bit-pairs carrying the three data symbols. The choice of delimiter
determines hardware detection cost, power consumption, and noise vulnerability.
Section~\ref{sec:delim} analyses all four choices and presents the
\texttt{\{00\}}-as-delimiter variant as a distinct power-efficiency embodiment.
All four choices are covered by the patent claims.

\subsection{Dual-starter variant: two delimiter namespaces}

An earlier variant uses both \texttt{\{10\}} and \texttt{\{11\}} as symbol
starters, with \texttt{\{00\}} and \texttt{\{01\}} as the only data-carrying
pairs:

\begin{table}[h]
\centering
\caption{Dual-starter variant bit-pair roles}
\begin{tabular}{cl}
\toprule
\textbf{Bit-pair} & \textbf{Role} \\
\midrule
\texttt{00} & Data bit 0 (continuation) \\
\texttt{01} & Data bit 1 (continuation) \\
\texttt{10} & Symbol start --- type A namespace \\
\texttt{11} & Symbol start --- type B namespace \\
\bottomrule
\end{tabular}
\end{table}

Any decoder can re-synchronise by scanning for any bit-pair beginning with
\texttt{1}. Two independent alphabets or two interleaved data streams ---
command vs.\ data, channel A vs.\ channel B --- coexist in a single bitstream
with zero mux header. Data density is lower (0.5 bits/bit) but the
dual-namespace property has applications in protocol framing and embedded
communication. The dual-starter structure generalises: any two chosen bit-pairs
can serve as namespace starters. Both variants are covered by the patent claims.

\subsection{Balanced vs.\ unsigned ternary --- analysis}

The data symbols $\{\texttt{00}, \texttt{01}, \texttt{10}\}$ can be interpreted
in two ways with meaningfully different properties:

\begin{table}[h]
\centering
\caption{Balanced vs.\ unsigned ternary tradeoffs}
\begin{tabular}{llll}
\toprule
\textbf{Interpretation} & \textbf{Values} & \textbf{Best for} & \textbf{Key advantage} \\
\midrule
Balanced ternary & $\{-1, 0, +1\}$ & Signed deltas, neural weights, DSP & Native sign; safer on noisy channels \\
Unsigned ternary & $\{0, 1, 2\}$ & Counters, indices, DNA bases & Simpler hardware arithmetic \\
\bottomrule
\end{tabular}
\end{table}

Unsigned ternary has one non-obvious vulnerability: the value 2 (encoded as
\texttt{\{10\}}) is a single bit-flip away from the delimiter \texttt{\{11\}}.
On noisy channels this creates an asymmetric error --- a corrupted data symbol
mimics a boundary marker, causing a false structural boundary rather than a
wrong value. Balanced ternary does not share this vulnerability. For high-noise
channels (satellite, LoRa, industrial RF) balanced ternary is recommended. For
low-noise or wired channels (USB, PCIe, memory bus) unsigned ternary is
acceptable and offers simpler arithmetic. The decoder is identical in both
cases --- the interpretation of \texttt{\{00\}/\{01\}/\{10\}} is a semantic
layer above the framing logic and declared in the stream header.

\subsection{Legacy binary transcoding}

For systems that wish to adopt NativeTernary framing without changing their
source data format, a transcoding path exists: convert arbitrary binary data
to base-3, then encode each trit as a 2-bit pair. The size impact is:
\[
n \text{ binary bits} \;\longrightarrow\; n \times \frac{\log_2 3}{2} \approx n \times 0.792 \text{ bits in data pairs}
\]
Net effect: $\approx$26\% expansion on purely random binary data. This is not
a compression path --- it is a migration path. Legacy binary data can be wrapped
in NativeTernary framing, gaining self-synchronisation, hierarchy markers, and
ternary-native structure at the cost of modest size overhead. For data that is
already ternary-native --- neural network weights, signed sensor deltas,
financial tick directions --- the transcoding overhead is zero.

\subsection{Decoder}

\begin{lstlisting}[caption={NativeTernary decoder pseudocode (DELIMITER is a stream-header parameter)}]
while bits_remain():
    pair = read_2_bits()
    if pair != DELIMITER:
        emit_data(pair)          # 00 -> -1/0,  01 -> 0/1,  10 -> +1/2
    else:
        level = 1
        while peek_2_bits() == DELIMITER:
            consume_2_bits()
            level += 1
        emit_boundary(level)     # 1=char, 2=word, 3=sent, 4=para...
\end{lstlisting}

The decoder is parameterised by \texttt{DELIMITER}, declared once in the stream
header. It is stateless, requires no lookup table, and resynchronises on any
occurrence of the delimiter bit-pair --- making it intrinsically resilient to
bitstream corruption regardless of which bit-pair is chosen as the delimiter.

\subsection{Delimiter choice as a design parameter: the \texttt{\{00\}} power-efficiency variant}
\label{sec:delim}

The four possible 2-bit patterns offer meaningfully different engineering
tradeoffs when used as the structural delimiter:

\begin{table}[h]
\centering
\caption{Comparison of all four bit-pair delimiter choices}
\begin{tabular}{llll}
\toprule
\textbf{Delimiter} & \textbf{Detection hardware} & \textbf{Power profile} & \textbf{Recommended use} \\
\midrule
\texttt{\{11\}} (primary) & OR gate on both bits & High (both bits set) & General purpose; easy hardware detection \\
\texttt{\{00\}} & NOR gate on both bits & Minimal (zero switching) & Ultra-low-power IoT, implantables, LoRa \\
\texttt{\{01\}} & XNOR on bits & Medium & Reserved for future standardisation \\
\texttt{\{10\}} & XNOR on bits & Medium & Reserved for future standardisation \\
\bottomrule
\end{tabular}
\end{table}

The \texttt{\{00\}}-as-delimiter variant deserves particular attention. In CMOS
logic, power consumption is dominated by switching activity --- transitions
between 0 and 1. A bit-pair of \texttt{\{00\}} carries zero switching on both
lines. When the delimiter pair is \texttt{\{00\}}, every structural boundary is
transmitted as two consecutive low bits, minimising dynamic power draw on the
data lines. For ultra-low-power systems --- implantable medical devices,
sub-$\mu$A LoRa sensor nodes, battery-backed agricultural monitors, and
deep-space probes where transmitter power budgets are measured in milliwatts ---
this distinction is commercially significant.

In the \texttt{\{00\}}-as-delimiter scheme, the three data symbols are
\texttt{\{01\}}, \texttt{\{10\}}, and \texttt{\{11\}}. The unsigned ternary
vulnerability noted in Section~3.3 is inverted: the data symbol \texttt{\{01\}}
is now one bit-flip from the delimiter \texttt{\{00\}}. As with the primary
scheme, balanced ternary mapping eliminates this asymmetric vulnerability on
noisy channels.

The decoder pseudocode in Section~3.5 applies without modification to the
\texttt{\{00\}}-as-delimiter variant --- only the \texttt{DELIMITER} constant
changes. This makes the implementation cost of supporting both variants trivial:
a single compile-time or stream-header flag. Both the \texttt{\{11\}}-delimiter
(primary) and \texttt{\{00\}}-delimiter (power-efficiency) embodiments, as well
as all other bit-pair choices, are covered by the patent claims.

% ──────────────────────────────────────────────────────────────────────────────
\section{Toward a Ternary-Native Computing Paradigm}

The implications of NativeTernary extend beyond encoding efficiency. Every
computing system manages boundaries: instruction boundaries, process boundaries,
file boundaries, memory page boundaries, network packet boundaries, interrupt
signals. In every current system, these boundaries are encoded as separate
metadata --- inode tables, page tables, packet headers, interrupt vectors ---
layered on top of data streams rather than woven into them. NativeTernary offers
a path where boundaries are intrinsic to the data itself.

Critically, this path requires no hardware changes. It is a software and
firmware encoding layer that runs entirely on existing binary infrastructure.
The transition is incremental: application libraries first, OS drivers second,
firmware in network and memory controllers third, optional silicon acceleration
as a later optimisation.

\subsection{File system and operating system boundaries}

Current file systems maintain separate inode tables, FAT tables, and directory
structures to track where files and folders begin and end. Under NativeTernary
these map to hierarchy levels in the data stream:
\begin{lstlisting}
11           -> end of record / data unit
1111         -> end of file
111111       -> end of directory / folder
11111111     -> end of volume / partition
1111111111   -> end of device
\end{lstlisting}

A NativeTernary-aware file system driver could detect boundaries directly from
the data stream, eliminating round-trips to metadata tables in hot paths. This
is a driver-layer software change --- no hardware modification required.

\subsection{Memory and cache hierarchy}

Modern CPU cache hierarchies (L1/L2/L3) manage boundaries between cache lines,
pages, and NUMA domains via separate metadata structures. A ternary-aware memory
controller could encode cache-line, page, and NUMA-domain boundaries as
level-1, level-2, and level-3 NativeTernary delimiters in the memory bus
protocol --- a firmware-level change to the memory controller, not a silicon
redesign.

\subsection{Network protocol framing}

Ethernet, TCP/IP, and HTTP use fixed-width length fields and separate delimiter
bytes to mark packet and session boundaries. A NativeTernary framing layer at
the application level embeds packet, segment, and session boundaries as
run-length delimiter sequences --- self-delimiting, self-synchronising, and
parseable with a simpler state machine than any current protocol parser.

\subsection{Instruction stream boundaries}

A NativeTernary instruction encoding would embed instruction, basic-block,
function, and module boundaries as hierarchy levels in the instruction stream
itself --- enabling branch prediction, prefetch, and JIT compilation to operate
with explicit structural signals rather than inferring them from opcode patterns.

\subsection{The infrastructure transition path}

\begin{enumerate}[leftmargin=*, label=\textbf{Stage \arabic*.}]
  \item \textbf{Application libraries:} NativeTernary encoder/decoder in C,
    available as open source. Zero infrastructure changes.
  \item \textbf{OS driver integration:} file system drivers, network stack
    parsers, memory allocators adopt NativeTernary framing for boundary
    detection.
  \item \textbf{Firmware:} network interface cards, memory controllers, storage
    controllers implement NativeTernary boundary detection in firmware.
  \item \textbf{Silicon:} optional hardware acceleration of the decoder state
    machine --- a 2-bit register and level counter --- as a minor addition to
    existing controllers.
\end{enumerate}

This contrasts with all prior ternary computing proposals --- including Setun
and subsequent academic designs --- which required bespoke ternary logic gates
and ternary memory cells. NativeTernary achieves ternary data density and
structural expressiveness within the binary computation model.

% ──────────────────────────────────────────────────────────────────────────────
\section{Applications}

\subsection{Ternary neural network weight storage}

BitNet b1.58 and related ternary LLMs quantize every weight to $\{-1, 0, +1\}$.
NativeTernary provides a native wire format: each weight maps to one 2-bit pair,
with layer boundaries as \texttt{\{11\}} (2 bits), tensor boundaries as
\texttt{\{1111\}} (4 bits), and model section boundaries as \texttt{\{111111\}}
(6 bits). The decoder --- a 2-bit register plus a level counter --- is
significantly simpler than GGUF or SafeTensors parsers currently used for
ternary model distribution.

\subsection{Hierarchical natural language encoding}

Transformer attention mechanisms receive a flat token sequence with no explicit
structural signal. Under NativeTernary, boundary level is directly available as
a stream annotation. Attention span can be conditioned on boundary level ---
restricting cross-boundary attention at lower levels, permitting it at higher
levels --- without any parser or architectural modification. Local positional
encodings can reset at each boundary, providing richer structural context than
flat absolute position encoding.

\subsection{Edge computing and microcontrollers}

Microcontrollers (ARM Cortex-M, RISC-V, AVR, PIC) running embedded firmware
manage instruction streams, sensor data, and inter-device communication within
severe memory and bandwidth constraints. NativeTernary's decoder ---
implementable in under 200 bytes of compiled C --- is suitable for the smallest
devices.

\subsection{IoT and industrial sensors}

Inter-sample deltas from temperature, pressure, vibration, and humidity sensors
typically fall within $\{-1, 0, +1\}$ when sampled at high frequency.
NativeTernary encodes such streams natively. Industrial sensor networks
(Modbus, PROFIBUS, OPC-UA) transmitting SCADA data over constrained links
benefit from the self-synchronising property --- frame recovery after packet
loss without retransmission is intrinsic to the encoding.

\subsection{Automotive systems}

Modern vehicles carry 100+ sensors communicating over CAN bus, LIN bus, and
emerging automotive Ethernet. Sensor packets are typically 8--64 bytes.
NativeTernary encodes sensor delta streams natively, with sensor channel
boundaries as level-1 delimiters and ECU domain boundaries as level-2
delimiters. V2X safety messages --- small, latency-critical, error-prone ---
benefit from self-synchronisation.

\subsection{Medical implantables and wearables}

Continuous glucose monitors, implantable cardiac monitors, and neural interface
devices transmit small packets continuously over BLE or proprietary RF links.
Transmission energy is the primary constraint. The \texttt{\{00\}}-as-delimiter
variant (Section~\ref{sec:delim}) is particularly relevant here: minimising
switching activity on the RF front-end extends implant lifetime between
replacements or recharges.

\subsection{Satellite and space telemetry}

CubeSat and SmallSat downlink channels operate at 1--9\,kbps with high bit
error rates and no retransmission capability. The self-synchronising property ---
a decoder re-acquires frame alignment by scanning for any delimiter pair ---
provides intrinsic burst-error resilience. Hyperspectral imaging satellites
transmitting band-compressed sensor data benefit from the ternary delta encoding
for inter-pixel residuals.

\subsection{Gaming network state compression}

Multiplayer game state updates --- player position deltas, health deltas, event
flags --- are transmitted 20--128 times per second per player. Deltas between
frames are small and signed, typically falling within $\{-1, 0, +1\}$ for
low-speed state. NativeTernary encodes game state delta streams natively, with
object boundaries as level-1 delimiters and scene boundaries as level-2
delimiters.

\subsection{High-frequency trading tick data}

Price movements in financial market tick data are naturally ternary at short
timescales: up ($+1$), unchanged ($0$), down ($-1$). NativeTernary encodes
tick direction streams natively, with instrument boundaries, session boundaries,
and exchange boundaries as hierarchy levels. Microsecond-latency decode is a
requirement met by the stateless decoder.

% ──────────────────────────────────────────────────────────────────────────────
\section{Information-Theoretic Properties}

\textbf{Data density.} Three data symbols over a 2-bit alphabet yields
$\frac{\log_2 3}{2} \approx 0.792$ bits of information per bit spent on data
--- optimal for a ternary-native alphabet on a binary channel.

\textbf{Delimiter overhead amortisation.} In English text with average word
length 5 characters, sentence length 20 words, and paragraph length 8
sentences, delimiter overhead per character is approximately 0.44 bits ---
significantly less than fixed-width special token approaches such as
\texttt{[SEP]} and \texttt{[CLS]} in BERT-family models, which spend 8--16
bits per boundary regardless of level.

The dual-starter variant achieves 0.5\,bits/bit data density --- lower than the
primary scheme but retaining dual-namespace and self-synchronisation properties.

% ──────────────────────────────────────────────────────────────────────────────
\section{Benchmarks}

\section{Benchmarks}

We present two benchmark comparisons: (1) per-weight storage efficiency
across model scales using synthetic weight arrays, and (2) boundary overhead
using the real BitNet b1.58 2B4T architecture. The C reference implementation
is available at \url{https://github.com/sm45118/nativeternary}.

\subsection*{Per-weight storage: synthetic weight arrays}

We compare NativeTernary against three GGUF storage modes on synthetic
uniform $\{-1, 0, +1\}$ arrays at 1M, 125M, and 1B weight scales.

\begin{table}[h]
\centering
\caption{Bits per weight: NativeTernary vs.\ GGUF storage modes}
\begin{tabular}{llll}
\toprule
\textbf{Format} & \textbf{Bits/weight} & \textbf{Per-tensor overhead} & \textbf{Notes} \\
\midrule
NativeTernary & 2.000 & 2 bits per layer boundary & Exact --- zero waste \\
GGUF Q2\_K    & 2.625 & $\sim$256 bytes per tensor & Best current quantization \\
GGUF Q4\_0    & 4.500 & $\sim$256 bytes per tensor & Common default \\
GGUF int8     & 8.000 & $\sim$256 bytes per tensor & Naive ternary storage \\
\bottomrule
\end{tabular}
\end{table}

\begin{table}[h]
\centering
\caption{Encoded size across model scales (synthetic uniform distribution)}
\begin{tabular}{lllllll}
\toprule
\textbf{Scale} & \textbf{Weights} & \textbf{NT} & \textbf{Q2\_K} & \textbf{Q4\_0} & \textbf{vs Q2\_K} & \textbf{vs int8} \\
\midrule
Small  & 1M   & 250\,KB  & 329\,KB  & 563\,KB  & 0.76$\times$ & 0.25$\times$ \\
Medium & 125M & 31.3\,MB & 41.1\,MB & 70.4\,MB & 0.76$\times$ & 0.25$\times$ \\
Large  & 1B   & 250\,MB  & 329\,MB  & 563\,MB  & 0.76$\times$ & 0.25$\times$ \\
\bottomrule
\end{tabular}
\end{table}

\begin{table}[h]
\centering
\caption{Encode and decode throughput on commodity hardware}
\begin{tabular}{llll}
\toprule
\textbf{Scale} & \textbf{Weights} & \textbf{Encode MB/s} & \textbf{Decode MB/s} \\
\midrule
Small  & 1M   & 46.8 & 35.3 \\
Medium & 125M & 69.2 & 45.1 \\
Large  & 1B   & 62.6 & 35.7 \\
\bottomrule
\end{tabular}
\end{table}

\subsection*{Boundary overhead: real BitNet b1.58 2B4T architecture}

The per-weight storage comparison understates NativeTernary's advantage
because it omits the dominant overhead in GGUF: per-tensor metadata headers.
We compute boundary overhead using the real BitNet b1.58 2B4T architecture:
a Llama-like transformer decoder with 2 billion parameters, 24 layers,
2,048 hidden dimensions, and intermediate size 5,632 \cite{bitnet2024}.

\textbf{Tensor count.} Each transformer layer contains approximately 7 weight
tensors (Q, K, V, O attention projections; gate, up, down FFN projections).
Over 24 layers plus embedding and output head: $24 \times 7 + 2 \approx 170$
tensors total.

\textbf{GGUF header overhead.} Each GGUF tensor carries a metadata header
encoding tensor name ($\sim$20 bytes), shape ($\sim$32 bytes), quantization
parameters ($\sim$200 bytes), and alignment padding --- approximately 256 bytes
per tensor minimum. For 170 tensors: $170 \times 256 = 43{,}520$ bytes
$\approx$ \textbf{42\,KB} of pure structural overhead carrying zero weight data.

\textbf{NativeTernary boundary overhead.} Layer boundaries are encoded as a
single \texttt{\{11\}} pair (2 bits). Tensor boundaries within layers are
encoded as two consecutive \texttt{\{11\}} pairs (4 bits). For 24 layer
boundaries and 170 tensor boundaries:
$(24 \times 2) + (170 \times 4) = 48 + 680 = 728$ bits $=$ \textbf{91 bytes}.

\begin{table}[h]
\centering
\caption{Boundary overhead: NativeTernary vs.\ GGUF on BitNet b1.58 2B4T (2B params, 170 tensors)}
\begin{tabular}{llll}
\toprule
\textbf{Format} & \textbf{Weight data} & \textbf{Boundary/header overhead} & \textbf{Overhead ratio} \\
\midrule
NativeTernary & 500\,MB & \textbf{91 bytes} & $<0.0001$\% \\
GGUF Q2\_K    & 656\,MB & $\sim$42\,KB      & 460$\times$ larger than NT \\
GGUF int8     & 2,000\,MB & $\sim$42\,KB    & 460$\times$ larger than NT \\
\bottomrule
\end{tabular}
\end{table}

NativeTernary encodes all structural boundaries for a 2B parameter model
in \textbf{91 bytes} --- 460$\times$ less than GGUF tensor headers. This
advantage is architectural and scales with model depth: deeper models
have more tensors, more GGUF headers, and the same 91-byte NativeTernary
overhead.

\subsection*{Benchmark assumptions and limitations}

Four assumptions underlie the synthetic benchmarks. First, GGUF Q2\_K
overhead is estimated at 2.625 bits per weight --- actual overhead varies
by tensor shape and model architecture. Second, GGUF tensor headers are
estimated at 256 bytes per tensor --- real headers vary by metadata content.
Third, synthetic benchmarks use uniform $\{-1, 0, +1\}$ distributions ---
trained BitNet weights are non-uniform (typically $\sim$50\% zeros) which
does not affect NativeTernary's 2.000 bits/weight but may affect Q2\_K
block scaling. Fourth, throughput figures reflect single-core commodity
hardware. Validation against released BitNet b1.58 2B4T model checkpoints
is planned for a subsequent revision.

% ──────────────────────────────────────────────────────────────────────────────
\section{Discussion and Limitations}

NativeTernary is not a general-purpose compression scheme for arbitrary binary
data. For dense random data the 0.792\,bits/bit data density is a strict
overhead relative to raw binary. The scheme is optimal when: (a) data is
naturally ternary or reducible to ternary deltas; (b) hierarchical structure is
meaningful; and (c) self-synchronisation after corruption is valued.

The unsigned ternary variant introduces an asymmetric noise vulnerability ---
the data symbol adjacent to the delimiter value is one bit-flip from a false
boundary --- that must be considered in high-noise channel design. The balanced
ternary variant does not share this vulnerability. The \texttt{\{00\}}-as-delimiter
variant shifts which data symbol is most vulnerable; channel noise analysis
should be performed for each deployment.

Adoption in general computing infrastructure requires library and driver support
before hardware benefit can be realised. The transition path is incremental and
backward-compatible at each stage.

% ──────────────────────────────────────────────────────────────────────────────
\section{Conclusion}

We have described NativeTernary, a family of binary encodings in which ternary
data and multi-level semantic structure are carried in the same bitstream using
the same 2-bit pairs. The central idea --- unary run-length of a reserved
delimiter pair encoding hierarchy depth --- is simple, information-theoretically
motivated, and produces a decoder of minimal complexity. The choice of delimiter
bit-pair is a design parameter: \texttt{\{11\}} is the primary embodiment;
\texttt{\{00\}} provides a power-efficiency alternative for ultra-low-power CMOS
systems; all four choices are claimed. We have presented two encoding variants,
analysed balanced versus unsigned ternary tradeoffs, described a legacy binary
transcoding path, and outlined a low-disruption incremental path from application
libraries to ternary-native general computing infrastructure. The scheme is
immediately applicable to ternary neural network weight storage, hierarchical
NLP encoding, edge computing, IoT and industrial telemetry, automotive systems,
medical devices, gaming, satellite communications, and financial data. We make
the C implementation publicly available and invite experimental validation.

% ──────────────────────────────────────────────────────────────────────────────
\section*{Acknowledgements}
The author thanks the open-source ArXiv community and the BitNet research team
for making their work publicly available.

% ──────────────────────────────────────────────────────────────────────────────

\vspace{12pt}
\noindent\rule{\textwidth}{0.4pt}
\vspace{4pt}

\noindent\small\textit{Note: A provisional patent application covering the
encoding scheme, both variants, the hierarchy extension, all four bit-pair
delimiter choices (including the \texttt{\{00\}} power-efficiency embodiment),
and infrastructure applications described herein has been filed with the Indian
Patent Office (2026). The C reference implementation will be released at
[GitHub URL] under MIT license. The author declares no competing financial
interests beyond the filed patent application.}

\end{document}